\documentclass[10pt,twocolumn,letterpaper]{article}

\usepackage{iccv}
\usepackage{times}
\usepackage{epsfig}
\usepackage{graphicx}
\usepackage{amsmath}
\usepackage{amssymb}
\usepackage{caption}
\usepackage{pifont}
\usepackage{xcolor}
\usepackage{marvosym}
\newcommand{\cmark}{\ding{51}}%
\newcommand{\xmark}{\text{\ding{55}}}

\usepackage[breaklinks=true,bookmarks=false]{hyperref}

\iccvfinalcopy 

\def\iccvPaperID{****} 

\ificcvfinal\fi

\begin{document}

\title{Motion-I2V: Consistent and Controllable Image-to-Video Generation with Explicit Motion Modeling}

\author{
Xiaoyu Shi$^{1*}$ \and
Zhaoyang Huang$^{7*}\textsuperscript{\Letter}$ \and
Fu-Yun Wang$^{1*}$ \and
Weikang Bian$^{1*}$ \and
Dasong Li$^{1}$ \and 
Yi Zhang$^{3}$ \and 
Manyuan Zhang$^{1}$ \and
Ka Chun Cheung$^{2}$ \and
Simon See$^{2}$ \and
Hongwei Qin$^{3}$ \and
Jifeng Dai$^{4}$ \and
Hongsheng Li$^{1,5,6}\textsuperscript{\Letter}$ \\ \and
$^{1}$Multimedia Laboratory, The Chinese University of Hong Kong \and
$^{2}$NVIDIA AI Technology Center \and
$^{3}$SenseTime Research \and
$^{4}$Tsinghua University \and
$^{5}$Centre for Perceptual and Interactive Intelligence (CPII) \and
$^{6}$Shanghai AI Laboratory \and
$^{7}$Avolution AI \and
\\ \centerline{xiaoyushi@link.cuhk.edu.hk}
}

\twocolumn[{
\maketitle
\renewcommand\twocolumn[1][]{#1}
\begin{center}
    \centering
    \includegraphics[trim={10cm 3cm 12cm 9.5cm}, clip, width=\textwidth]{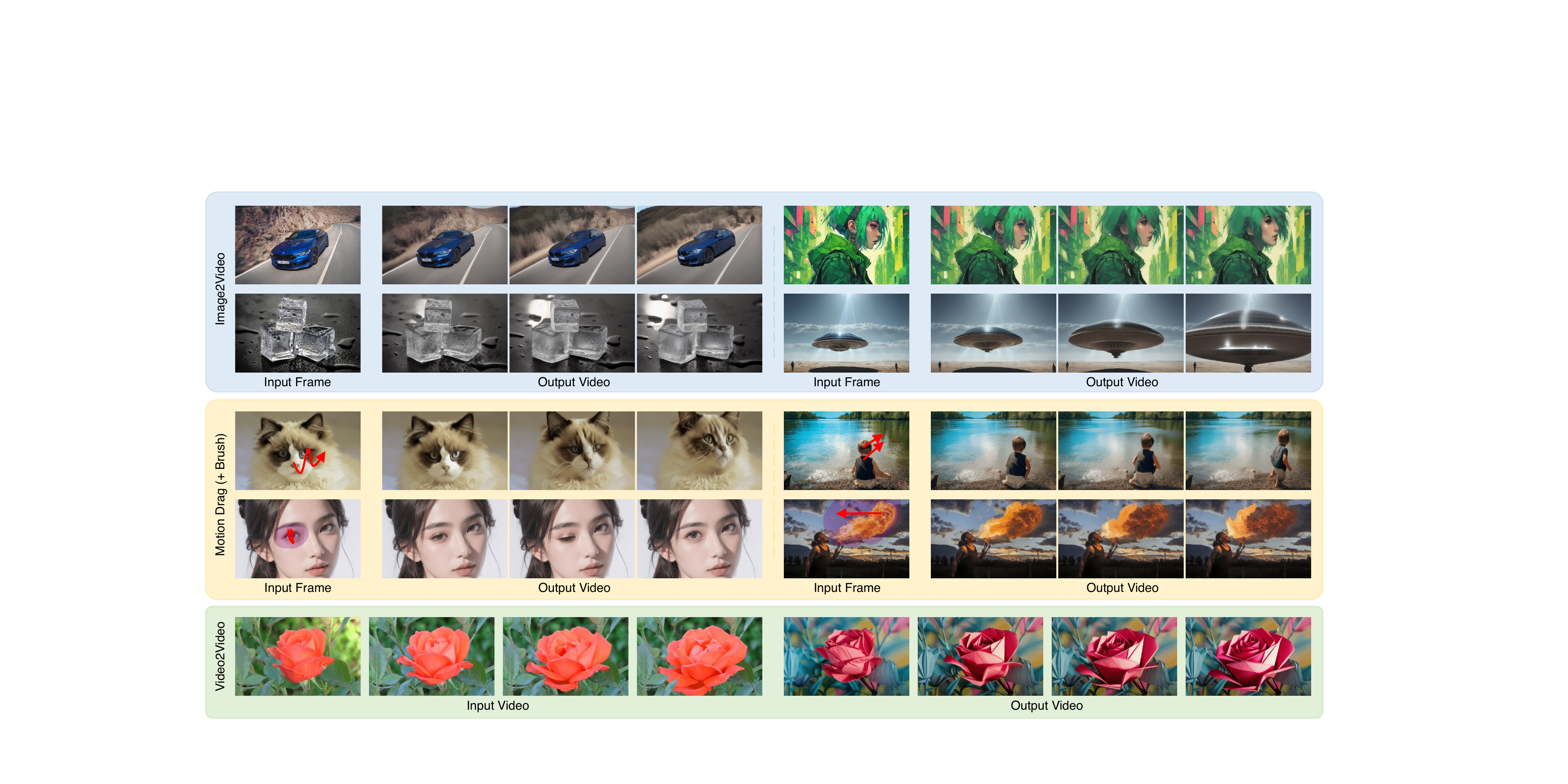}
    \captionof{figure}{
    \textit{\textbf{Motion-I2V} can generate consistent image-to-video results with large motion and viewpoint change. It also naturally supports users to more precisely control the motion trajectories and animated region with sparse trajectories (\textcolor{red}{red curved arrow}) and motion brush (\textcolor{violet}{purple mask}). Additionally, the second stage of Motion-I2V also supports zero-shot video-to-video translation.
    }}
    \label{fig:teaser}
\end{center}
}]


\begin{abstract}
   We introduce Motion-I2V, a novel framework for consistent and controllable image-to-video generation (I2V). In contrast to previous methods that directly learn the complicated image-to-video mapping, Motion-I2V factorizes I2V into two stages with explicit motion modeling. For the first stage, we propose a diffusion-based motion field predictor, which focuses on deducing the trajectories of the reference image's pixels. For the second stage, we propose motion-augmented temporal attention to enhance the limited 1-D temporal attention in video latent diffusion models. This module can effectively propagate reference image's feature to synthesized frames with the guidance of predicted trajectories from the first stage. Compared with existing methods, Motion-I2V can generate more consistent videos even at the presence of large motion and viewpoint variation. By training a sparse trajectory ControlNet for the first stage, Motion-I2V can support users to precisely control motion trajectories and motion regions with sparse trajectory and region annotations. This offers more controllability of the I2V process than solely relying on textual instructions. Additionally, Motion-I2V's second stage naturally supports zero-shot video-to-video translation. Both qualitative and quantitative comparisons demonstrate the advantages of Motion-I2V over prior approaches in consistent and controllable image-to-video generation. Please see our project page at \url{https://xiaoyushi97.github.io/Motion-I2V/}.
\end{abstract}

\section{Introduction}

Image-to-video generation (I2V) targets at animating a given image to a video clip with natural dynamics, while preserving the visual appearance. It has widespread applications in fields of film industry, augmented reality, automatic advertising and content creation for social media platforms. Traditional I2V methods, however, focus on specific categories (e.g. human hair~\cite{xiao2023automatic}, fluid~\cite{Holynski_2021_CVPR,mahapatra2021controllable,mahapatra2023synthesizing,Okabe2009AnimatingPO}, portraits~\cite{10.1145/3272127.3275043,WANG_2020_WACV,wang2020ImaGINator,wang2022latent,geng_facialanimation}). Consequently, such specialization restricts their utility in more diverse, open-domain scenarios.

In recent years, diffusion models~\cite{rombach2021highresolution,Imagen,glide} trained on web-scale image datasets have made impressive strides in producing high-quality and diverse images. Encouraged by this success, researchers have begun extending these models to the realm of I2V, aiming to leverage the strong image generative priors. These works~\cite{zhang2023pia,dai2023finegrained,xing2023dynamicrafter,2023videocomposer,2023i2vgenxl} typically equip text-to-image (T2I) models with 1-D temporal attention modules to create video base models. However, I2V presents more challenges compared to static image generation. It requires modeling the complicated spatial-temporal priors. The narrow temporal receptive field of 1-D temporal attention makes it difficult to ensure temporal consistency of the generated videos, especially in the presence of large motion. Another notable shortage of current I2V works is their limited controllability. These models primarily utilize the reference image and textual instructions as the generation conditions, but lack precise and even interactive control of the generated motions. This is in stark contrast to the field of image manipulation, where techniques like drag-based~\cite{shi2023dragdiffusion,mou2023dragondiffusion,10.1145/3588432.3591500} and region-specific~\cite{huang2023region,where2edit} controls have demonstrated substantial efficacy.

To remedy the aforementioned issues, we present Motion-I2V, a framework that factorizes image-to-video generation into two stages. The first stage focuses on predicting the plausible motions, in the form of pixel-wise trajectories. With such explicit motion modeling, the second stage is responsible for generating consistent animation with the predicted dynamics from the first stage. Specifically, in the first stage, we tune a pre-trained video diffusion model for motion field prediction. It takes the reference image and textual instruction as conditions, and predicts the trajectories of all pixels in the reference image. In the second stage, we propose a motion-augmented temporal attention to enhance the video diffusion model. The latent features of the reference image are warped according to all pixels' predicted trajectories and act as guidance via adaptively (through cross-attention) injecting into the synthesized frames at multiple scales. This warping operation brings dynamic temporal receptive field and alleviates the pressure of learning the complicated spatial-temporal patterns with only 1-D temporal attention.

Inspired by previous successes of adapting pre-trained large-scale model~\cite{gao2023clip,zhang2021tip,fang2023feataug,fang2023instructseq}, we also train a ControlNet~\cite{zhang2023controlnet} for motion prediction in the first stage, which takes sparse trajectories as the condition and generates plausible dense trajectories. This design empowers users to manipulate object motions with very sparse trajectory annotations. Our framework also naturally supports region-specific animation (named {\it motion brush}), enabling users to animate selected image areas with custom motion masks. Moreover, the second stage of Motion-I2V is capable of achieving video-to-video translation, where the trajectories are obtained from the source video. Users can transform the first frame with existing image-to-image tools and consistently propagate the transformed first frame using the second stage of Motion-I2V. These characteristics grant users enhanced controllability over the I2V process.

\begin{figure*}[!t]
  \includegraphics[trim={8cm 6cm 9cm 6cm}, clip, width=\textwidth]{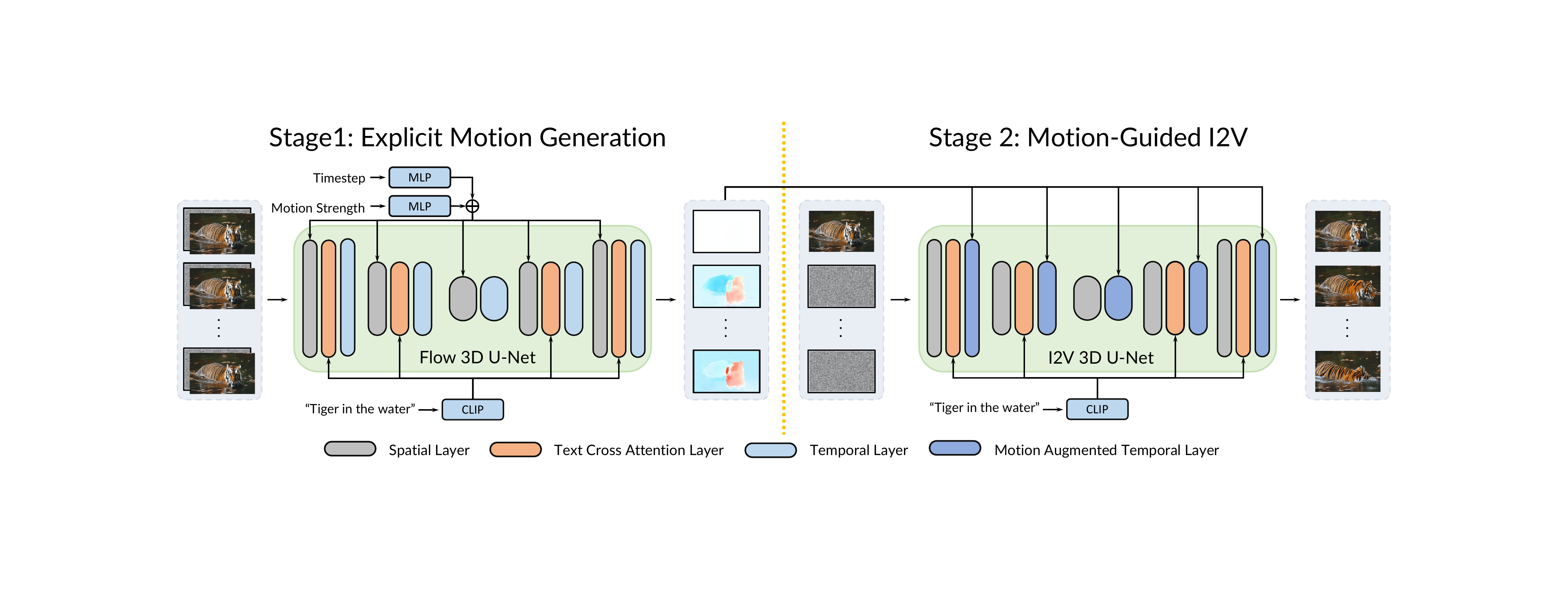}
  \caption{\textbf{Overview of Motion-I2V.} The first stage of Motion-I2V targets at deducing the motions that can plausibly animate the
reference image. It is conditioned on the reference image and text prompt, and predicts the motion field maps between the reference frame and all the future frames. The second stage propagates reference image's content to synthesize frames. A novel motion-augmented temporal layer enhances 1-D temporal attention with warped features. This operation enlarges the temporal receptive field and alleviates the complexity of directly learning the complicated spatial-temporal patterns.}
  \label{fig:overview}
\end{figure*}

\section{Related Work}

\subsection{Image Animation} Animating a single image has attracted a lot of attention in the research field. Previous approaches simulate motion for natural dynamics \cite{Holynski_2021_CVPR, li2023_3dcinemagraphy, mahapatra2021controllable, Okabe2009AnimatingPO,Xiong_2018_CVPR,rottshaham2019singan,Animating_cloud}, human faces \cite{wang2020ImaGINator,wang2022latent,geng_facialanimation} and bodies \cite{Weng2018PhotoW3,wang2022latent,siarohin2021motion,karras2023dreampose,Blattmann2021Understanding}. Some of the previous methods employ optical flow to model the motion and uses warping-based rendering techniques. We get inspiration from this line of research and introduce explicit motion modeling into modern generative models. Recent developments on image animation are driven by diffusion models \cite{rombach2021highresolution,ho2020denoising,song2021denoising}. Mahapatra \textit{et al.} \cite{mahapatra2023synthesizing} transplant the estimated optical flow to artistic paintings with a pre-trained text-to-image diffusion model. Li \textit{et al.} \cite{li2023generative} utilize a diffusion model to handle the natural oscillating motions. These animation approaches can only synthesize specific types of content and motion such as time-lapse videos and body animation. To solve this problem, some diffusion-based methods~\cite{xing2023dynamicrafter,2023i2vgenxl,2023videocomposer,zhang2023pia,dai2023finegrained}, are proposed to address the challenge of open-domain image animation. They capitalize on the strong generative priors of pre-trained diffusion models and have achieved unprecedented open-domain animation performance. However, these typically rely on vanilla 1-D temporal attention to learn the complicated image to video mapping. We propose to enlarge the receptive fields with explicit motion prediction.
\subsection{Diffusion Models}
Diffusion models (DMs) \cite{ho2020denoising,song2021denoising} have recently shown more stable training, better sample quality, and flexibility than VAE \cite{VAE}, GAN \cite{GAN} and FLow models \cite{flow_models}. DALL-E 2\cite{DALL-E}, GLIDE \cite{glide} and Imagen \cite{Imagen} employ diffusion models for text-to-image generation by conducting the diffusion process in pixel space, guided by language models \cite{Clip,T5} or classifier-free approaches. Stable diffusion \cite{rombach2021highresolution} shows unprecendented power on text-to-image generation by performing denoising diffusion on the latent space and supports many downstream applications~\cite{zhang2023unified,zhang2023kbnet}.
\newline
\noindent Recent attention has also been paid to employing diffusion models \cite{rombach2021highresolution} for video synthesis. 
Notably, ImagenVideo \cite{Ho2022ImagenVH} and Make-A-Video \cite{singer2022makeavideo} perform denoising diffusion in video pixel space, while MagicVideo \cite{zhou2023magicvideo} models the video distribution in the latent space. 
Video-P2P \cite{liu2023videop2p} and vid2vid-zero \cite{vid2vid-zero} propose to edit the video via cross-attention map manipulation. Text-to-video zero \cite{text2video-zero} construct latent code to model dynamics to employ stable diffusion models \cite{rombach2021highresolution} for video generation. Wang \textit{et al.} propose a versatile pipeline for extending the video generation length~\cite{wang2023gen}. VideoComposer \cite{2023videocomposer} adopts textual condition, spatial conditions and temporal conditions on the video diffusion models. Zhang \textit{et al.} propose a cascaded i2vgen-XL \cite{2023i2vgenxl} to ensure semantically and qualitatively excellent video generation. Dynamicrafter \cite{xing2023dynamicrafter} proposes a dual-stream image injection mechanism to utilize the motion prior of text-to-video diffusion models. These methods \cite{2023videocomposer, xing2023dynamicrafter, 2023i2vgenxl} usually allow the diffusion model to handle motion modeling and video generation simulateously, which leads to unrealistic motions and temporal inconsistent visual details. Our method decouples the motion modeling and video details generation to achieve realistic motions and preserve the pleasant details.

\subsection{Motion Modeling}
Motion modeling aims to understand and predict the movement of objects.
Optical flow is a common approach to represent motion, which estimates the displacement field between two consecutive frames. Early work formulated optical flow estimation as an optimization problem that utilized handcrafted features to maximize the visual similarity between image pairs~\cite{horn1981determining, black1993framework, bruhn2005lucas, sun2014quantitative}.
Deep learning-based methods have recently revolutionized the field of optical flow estimation. FlowNet~\cite{dosovitskiy2015flownet} was the first to introduce deep learning into end-to-end optical flow estimation, which demonstrated the potential of deep learning in this domain. After that, well-designed neural network architecture and synthetic datasets promoted the progress of optical flow estimation~\cite{ilg2017flownet, ranjan2017optical, sun2018pwc, sun2021loftr, hui2018liteflownet, hui2020lightweight, yang2019volumetric}. RAFT~\cite{teed2020raft} adopting iterative refinement with correlation volume significantly improved the performance. FlowFormer~\cite{huang2022flowformer,shi2023flowformer++} successfully applied the attention mechanism. Recently, VideoFlow~\cite{shi2023videoflow} explored the temporal information between multiple frames and achieved state-of-the-art accuracy.
%
Point tracking is another motion modeling method that computes the trajectory of the query point throughout the video frames~\cite{harley2022particle, doersch2022tap, zheng2023pointodyssey}. Context-PIPs~\cite{weikang2023context} and CoTracker~\cite{karaev2023cotracker} further improve the accuracy of point tracking with the context information. In recent work, DOT~\cite{moing2023dense} demonstrated that using sparse point tracing to initialize arbitrary distance optical flow estimation can maintain high accuracy with reasonable computational overhead.

\section{Method}

Given a reference image $I_0$ and a text prompt $c$, image-to-video synthesis (I2V) targets at generating a sequence of subsequent video frames $\{\hat{I}_1,\hat{I}_2,\dots,\hat{I}_N\}$. The key objective is to ensure that the generated video clip not only exhibits plausible motion but also faithfully preserves the visual appearance of the reference image. By leveraging the strong generative priors of diffusion models, recent methods have shown promising open-domain I2V generalization capacity. However, existing methods struggle to maintain temporal consistency largely due to the limited 1-D temporal attention mechanism. Meanwhile, they offer limited control over the generation results. In view of these limitations, we propose Motion-I2V, a novel framework that factorizes image-to-video generation into two stages, as shown in Fig.~\ref{fig:overview}. The first stage, detailed in Sec. \ref{stage1}, focuses on predicting plausible motions in the form of pixel-wise trajectories. Building on the predicted motion field, the second stage, described in Section \ref{stage2}, utilizes our proposed warpping-augmented temporal attention to synthesize future frames. We start with introducing the preliminary knowledge of the latent diffusion model~\cite{rombach2021highresolution} and video diffusion model in Sec. \ref{Preliminaries}.

\subsection{Preliminaries} \label{Preliminaries}
\noindent \textbf{Latent diffusion model.} We choose Latent Diffusion Model~\cite{rombach2021highresolution} (LDM) as the backbone generative model. It conducts the denoising process in the latent space of a Variational Autoencoder (VAE). During training, the input image $x_0$ is first encoded into a latent representation $z_0=\mathcal{E}(x_0)$ with the frozen encode $\mathcal{E}(\cdot)$. This latent code $z_0$ is then perturbed as:
\begin{equation}
    z_t = \sqrt{\overline{\alpha}_t} z_0 + \sqrt{1-\overline{\alpha}_t} \epsilon, \epsilon \sim \mathcal{N}(0,I),
\end{equation} \label{add-noise}
where $\overline{\alpha}_t = \prod_{i=1}^{t} (1-\beta_t)$ with $\beta_t$ is the noise strength coefficient at step $t$, and $t$ is uniformly sampled from the timestep index set $\{1,\dots,T\}$. This process can be regarded as a Markov chain, which incrementally adds Gaussian noise to the latent code $z_0$. The denoising model $\epsilon_\theta$ receives $z_t$ as input and is optimized to learn the latent space distribution with the objective function
\begin{equation}
    l_{\epsilon} = || \epsilon - \epsilon_{\theta}(z_t,t,c) ||_{2}^{2},
\end{equation}
where $c$ represents the condition, and is the user-provided text prompt in our case. In this paper, we choose Stable Diffusion 1.5 as the base LDM, where the denoising model $\epsilon_\theta$ is implemented as a U-Net architecture.

\noindent \textbf{Video latent diffusion model.}
\label{vldm}
We follow the previous works~\cite{guo2023animatediff,blattmann2023align,xing2023dynamicrafter} to expand the image LDM by incorporating temporal modules to create the video latent diffusion model (VLDM). Specifically, the spatial modules from the original image LDM are initialized with the pretrained weights and are frozen during training. This is to inherit the generative priors from the image LDM. Temporal modules $l^i_\phi$ comprise of efficient 1-D temporal attention for efficiency are inserted after each spatial attention block $l^i_\theta$. Given a 5-D video tensor $z$ of shape $batch \times channels \times frames \times height \times width$, passed through the temporal modules, its spatial dimensions $height$ and $width$ are reshaped to the batch dimension, yielding 1-D $(batch \times height \times width)$ feature sequences of length $frames$, and are transformed by the self-attention blocks. Such temporal modules are responsible for capturing the temporal dependencies between features of the same spatial location across different frames. For clarity and ease of discussion, we refer to such a VLDM variant with 1-D temporal attentions as the vanilla VLDM.

\subsection{Motion Prediction with Video Diffusion Models} \label{stage1}

The first stage of our proposed image-to-video generation framework targets at deducing the motions that can plausibly animate the reference image. As the latest large-scale diffusion models have been trained on web-scale text-image data, they contain rich knowledge of visual semantics. Such semantic knowledge can greatly benefit motion prediction once the model is trained to associate motion distributions with corresponding objects. Therefore, we choose to adapt the pre-trained stable diffusion model for video motion fields prediction to capitalize on the strong generative priors.

\noindent \textbf{Motion fields modeling.}
We denote the predicting target of the first stage, the motion fields that animate the reference image, as a sequence of 2D displacement maps $\{f_{0 \rightarrow i} | i=1,\dots,N\}$, where each $f_{0 \rightarrow i} \in \mathbb{R}^{2 \times H\times W}$ is the optical flow between the reference frame and future frame at timestep $i$. With such a motion fields representation, for each source pixel $\mathbf{p} \in \mathbb{I}^{2}$ of the reference image $I_0$, we can easily determine its corresponding coordinate $\mathbf{p}'_{i} = \mathbf{p} + f_{0 \rightarrow i}(\mathbf{p})$ on the target image $I_{i}$ at timestep $i$.

\noindent \textbf{Training a motion field predictor.} To learn a motion field prediction VLDM, we propose a three-step fine-tuning strategy. Initially, we tune a pre-trained LDM to predict a single displacement field conditioned on the reference image and text prompt. Subsequently, we freeze the tuned LDM parameters and integrate the vanilla temporal modules (as described in Sec. \ref{vldm}) to create an VLDM for training. This integration allows the model to learn the video's temporal motion distribution by jointly denoising the whole sequence of the motion fields. After training the temporal modules, we proceed to finetune the entire VLDM model to obtain the final motion fields predictor. We use FlowFormer++~\cite{shi2023flowformer++} and DOT~\cite{moing2023dense} to estimate optical flow and multi-frame trajectories as ground truth during training, respectively.

\noindent \textbf{Encoding motion fields and conditonal image.} As we choose the latent diffusion model for its computational efficiency, we encode each flow map $f_{0\rightarrow i} \in \mathbb{R}^{2 \times H\times W}$ into a latent representation $z_{0\rightarrow i,0}=\mathcal{E}_{flow}(f_i)\in \mathbb{R}^{4 \times h\times w}$ using an optical flow VAE encoder, where $h=H/8$ and $w=W/8$. The optical flow autoencoder mirrors the LDM image autoencoder's structure, except that it receives and outputs 2-channel optical flow map rather than 3-channel RGB images. To support image conditioning, we concatenate the latent code of clean reference image $\mathcal{E}(I_0)\in \mathbb{R}^{4 \times h\times w}$ along the channel dimension. We initialize all available LDM weights from the SD 1.5 checkpoint, and set weights for the newly added $4$ input channels to zero. Additionally, frame stride $i$ is embedded using a two-layer $MLP$ and is added to the time embeddings, serving as a motion strength condition.

\subsection{Video Rendering with Predicted Motion} \label{stage2}

The second stage of Motion-I2V targets at propagating the content of the reference image according to the predicted motion fields from stage 1 to synthesized frames, maintaining the fidelity and temporal consistency. We propose a motion-augmented temporal attention to enhance the vanilla 1-D temporal attention, guided by the predicted motion fields from the first stage. This operation enlarges the temporal receptive field and alleviates the pressure of directly predicting the complicated spatial-temporal patterns from a single image.

\noindent \textbf{Motion-augmented temporal attention.} We enhance vanilla VLDM's 1-D temporal attention with the proposed motion-augmented temporal attention and keep its other modules as is. Consider a latent feature $z\in \mathbb{R}^{(1+N) \times C_l \times h_l\times w_l}$ in the $l-th$ temporal layer $l^i_\phi$, where $c_l$, $h_l$, $w_l$ represent the channel dimension, height and width of the feature, respectively. We omit the batch dimension for brevity. Here we use $z[0] \in \mathbb{R}^{1 \times C_l \times h_l\times w_l}$ to denote the feature map corresponding to the reference frame, and $z[1:N] \in \mathbb{R}^{N \times C_l \times h_l\times w_l}$ for the subsequent frames. With the predicted motion fields $\{f_{0 \rightarrow i} | i=1,\dots,N\}$ (assuming resized to align the spatial shape) from the first stage, we forward-warp~\cite{niklaus2020softmax} $z[0]$ according to each of the motion field $f_{0\rightarrow i}$ as:
\begin{equation}
    z[i]'=\mathcal{W}(z[0], f_{0\rightarrow i}).
\end{equation}
These warped feature maps $z[i]'$ are interleaved with original feature maps along the temporal dimension to create augmented features $z_{aug}=[z[0], z[1]', z[1],...,z[N]', z[N]]\in \mathbb{R}^{(1+2\times N) \times C_l \times h_l\times w_l}$. Then $z$ and $z_{aug}$ are reshaped to $z' \in \mathbb{R}^{(h_l\times w_l) \times (1+N) \times C_l}$ and $z_{aug}'\in \mathbb{R}^{(h_l\times w_l) \times (1+2\times N) \times C_l}$, respectively. In other words, the spatial dimensions are shifted to batch axis and they are treated as 1-D tokens. The reshaped feature maps will be projected and go through the 1-D temporal attention layer:
\begin{equation}
    z'' = \rm{Attention}(Q,K,V)=\rm{Softmax}(QK^T)V,
\end{equation} \label{soft-inject}
where $Q=W^Qz'$, $K=W^Kz'_{aug}$ and $V=W^Vz'_{aug}$ are the three projections. Notably, the warped feature maps only serve as key and value features. Additionally, we add sinusoidal position encoding to $z$ and $z_{aug}$ to make the network aware of the temporal order the interleaved augmented feature maps. This operation enlarges the receptive fields of the temporal modules guided by the predicted motion fields from the first stage.

\noindent \textbf{Selective noising.} At each timestep $t$ of the denoising process, we always concatenate the $\it{clean}$ reference image's latent code $z_{ref} \in \mathbb{R}^{1 \times 4 \times h\times w}$ with other noisy latent codes $z_{0:N,t} \in \mathbb{R}^{N \times 4 \times h\times w}$ along the temporal axis. This guarantees that the reference image's content is faithfully preserved in the generation process.

\section{Fine-grained Control of Motion-I2V Generation}

Relying solely on textual prompt can lead to a lack of fine-grained control of the generation results. This limitation often results in users engaging in multiple rounds of trial and error to achieve their desired creation. In this section, we show that, by virtue of the explicit motion modeling, our Motion-I2V naturally supports fine-grained controls over the I2V process.

\subsection{Sparse Trajectory Guided I2V}
\begin{figure}[h]
  \includegraphics[trim={17cm 3.5cm 16cm 2cm}, clip, width=0.5\textwidth]{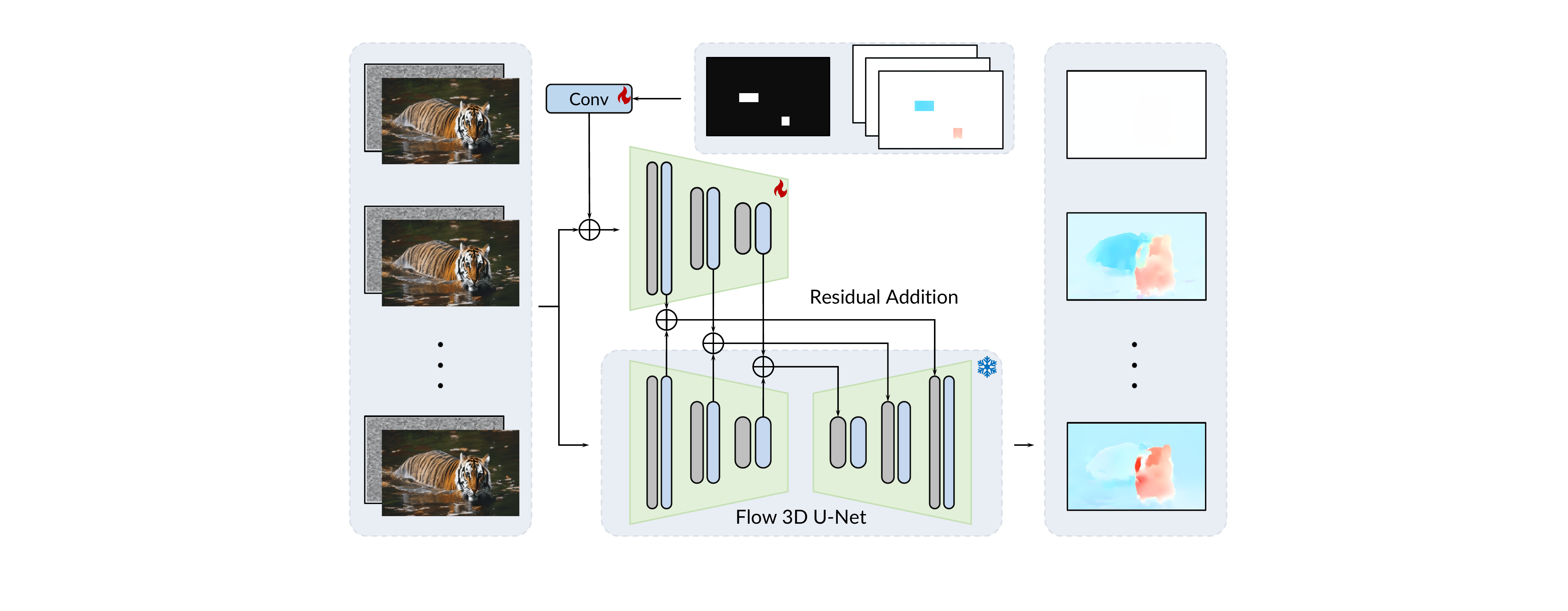}
  \caption{\textbf{Overview of trajectory ControlNet.} We train a Trajectory ControlNet based on the pre-trained stage 1 of Motion-I2V. It takes sparse trajectories and corresponding binary mask as additional conditions, and output dense optical flow maps.}
  \label{fig:controlnet}
\end{figure}
\begin{figure}[h]
    \small
    \setlength{\tabcolsep}{0.75pt}
    \begin{center}
    \begin{tabular}{@{} c c c c @{}}
        \includegraphics[width=.25\linewidth]{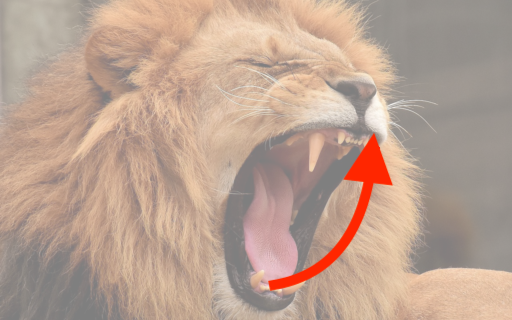} &
        \includegraphics[width=.25\linewidth]{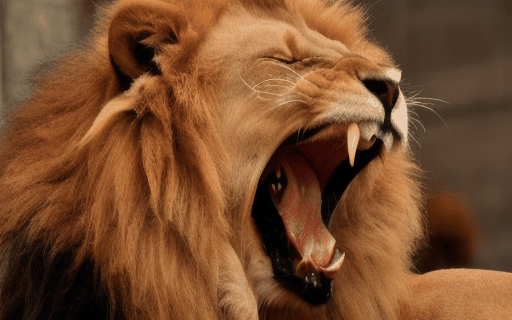} &
        \includegraphics[width=.25\linewidth]{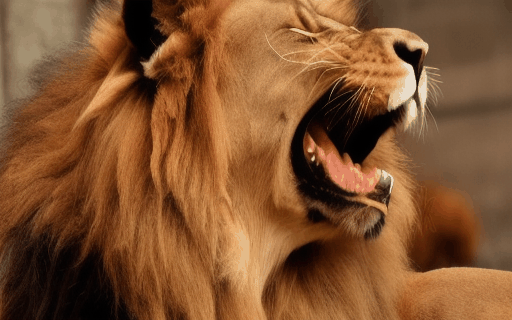} &
        \includegraphics[width=.25\linewidth]{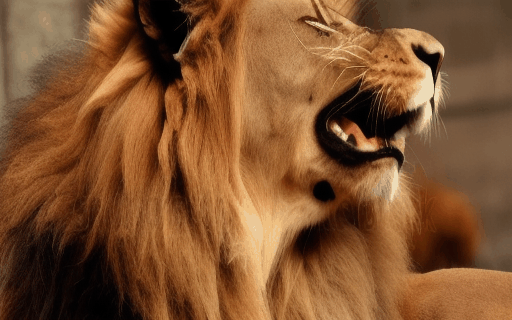}\\
        \includegraphics[width=.25\linewidth]{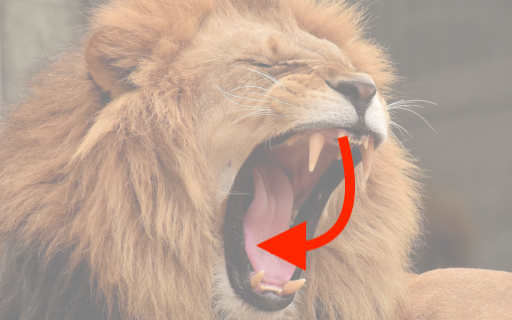} &
        \includegraphics[width=.25\linewidth]{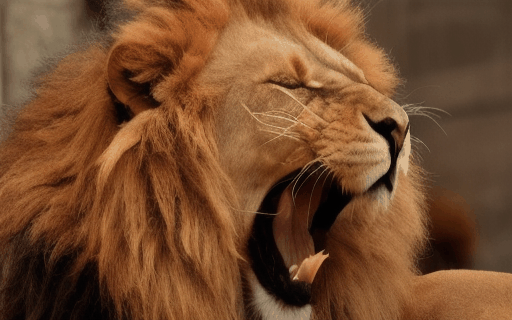} &
        \includegraphics[width=.25\linewidth]{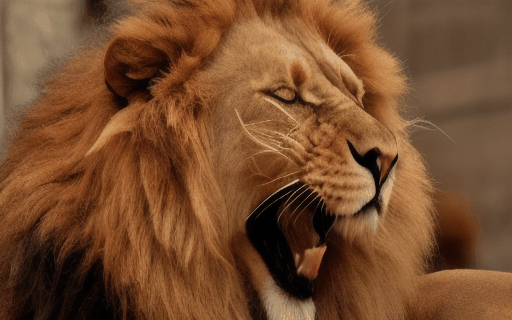} &
        \includegraphics[width=.25\linewidth]{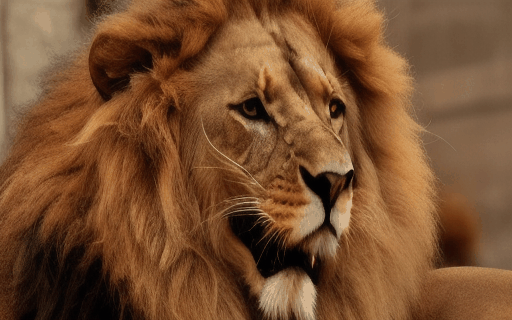}\\
        Input+Drag & Frame 2 & Frame 9 & Frame 16 \\
    \end{tabular}
    \end{center}
    \caption{\textbf{Examples of sparse trajectory guided I2V.} Users can precisely control the synthesized motions by drawing one or multiple trajectories (\textcolor{red}{red curved arrow}).}
    \label{fig:drag}
\end{figure}

We propose sparse trajectory guided I2V as an extension of our Motion-I2V framework. Specifically, given an image, users can draw one or multiple pixel-wise trajectories to precisely specify the desired motion of target pixels. Our network is designed to interpret these sparse trajectory inputs and transform them into plausible dense displacement fields with generative priors. Subsequently, these dense motion fields are utilized as inputs for the second stage of Motion-I2V. This strategy effectively enables users to interactively control the I2V process, as shown in Fig.~\ref{fig:drag}.

To achieve this intuitive setting, we train a ControlNet for the first stage, as shown in Fig.~\ref{fig:controlnet}. Specifically, we clone the down-sample and middle blocks of the 3D-Unet in the first stage as the ControlNet branch. This trainable ControlNet branch is connected to the frozen main branch with zero-initialized convolution layers following~\cite{zhang2023controlnet}. The ControlNet additionally takes sparse displacement fields $f_{sparse} \in \mathbb{R}^{N \times 2 \times H\times W}$ and a binary mask $m \in \{0,1\}^{H\times W}$ as conditions, where $1$ indicates pixels with given motion. We use a shallow 3D Conv network to encode the concatenation of $f_{sparse}$ and $m$ into 4-dim feature maps and add to the noisy latents as residuals. Please refer to Supplementary for training details.

\subsection{Region-Specific I2V}
\begin{figure}[h]
    \small
    \setlength{\tabcolsep}{0.75pt}
    \begin{center}
    \begin{tabular}{@{} c c c c @{}}
        \includegraphics[width=.25\linewidth]{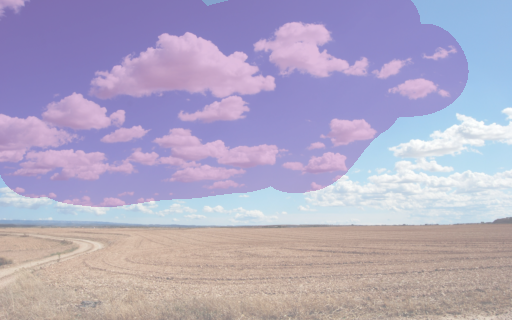} &
        \includegraphics[width=.25\linewidth]{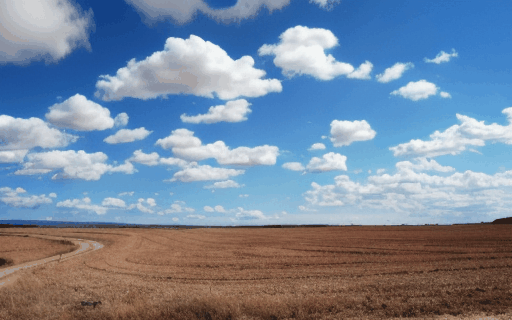} &
        \includegraphics[width=.25\linewidth]{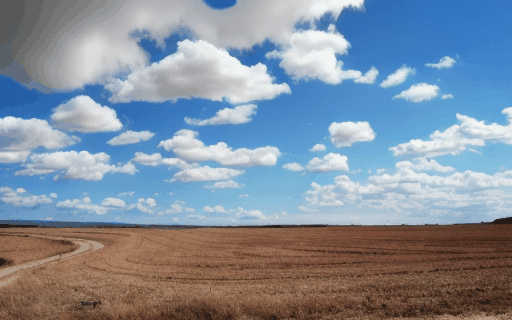} &
        \includegraphics[width=.25\linewidth]{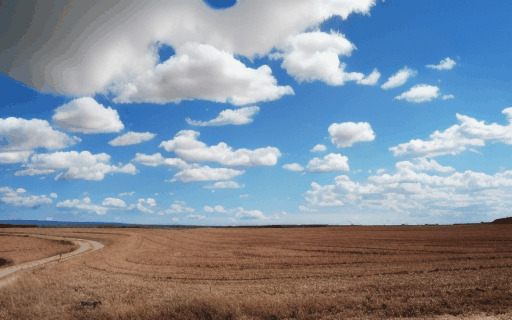}\\
        \includegraphics[width=.25\linewidth]{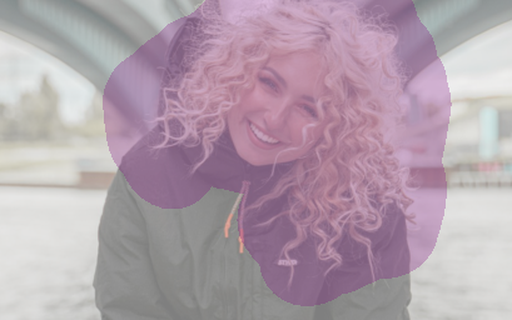} &
        \includegraphics[width=.25\linewidth]{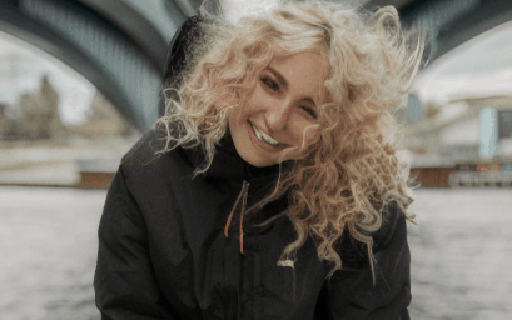} &
        \includegraphics[width=.25\linewidth]{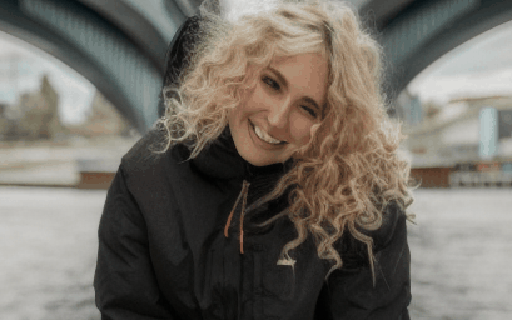} &
        \includegraphics[width=.25\linewidth]{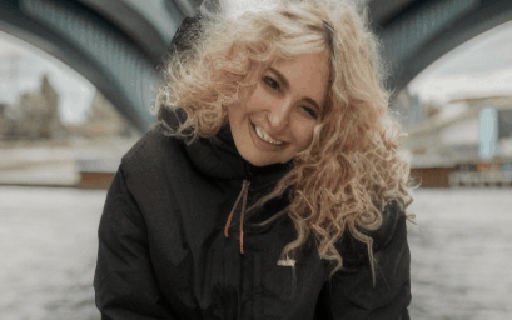}\\
        Input+Brush & Frame 2 & Frame 9 & Frame 16 \\
    \end{tabular}
    \end{center}
    \caption{\textbf{Examples of region-specific I2V.} Users can precisely Specify the animated regions by motion brush (\textcolor{violet}{purple mask}). Unmasked regions remains static.}
    \label{fig:motion-brush}
\end{figure}

\begin{figure}[h]
    \small
    \setlength{\tabcolsep}{0.75pt}
    \begin{center}
    \begin{tabular}{@{} c c c c @{}}
        \includegraphics[width=.25\linewidth]{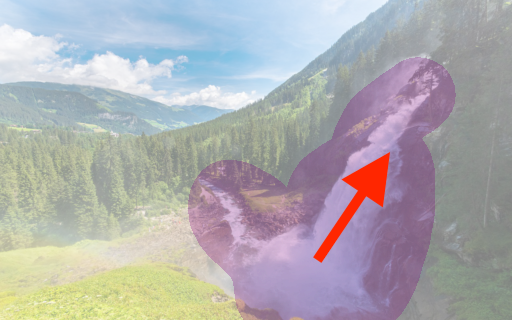} &
        \includegraphics[width=.25\linewidth]{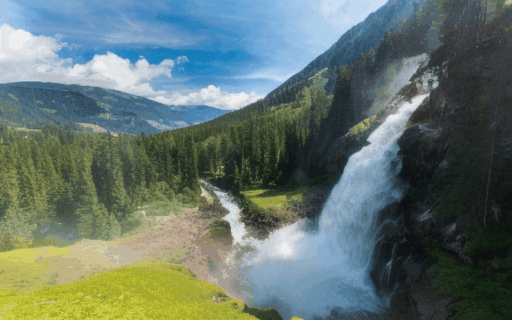} &
        \includegraphics[width=.25\linewidth]{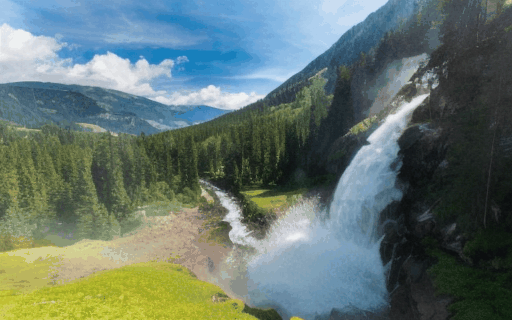} &
        \includegraphics[width=.25\linewidth]{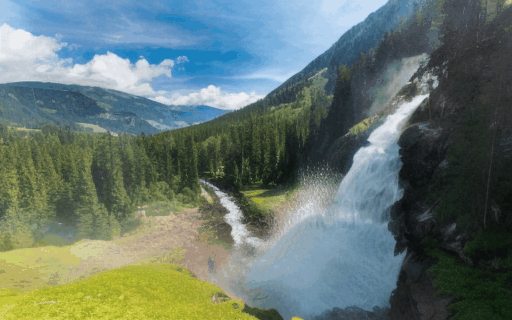}\\
        \includegraphics[width=.25\linewidth]{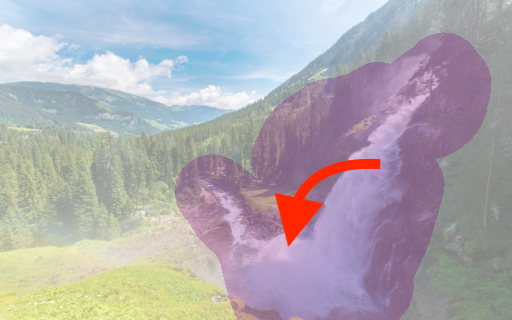} &
        \includegraphics[width=.25\linewidth]{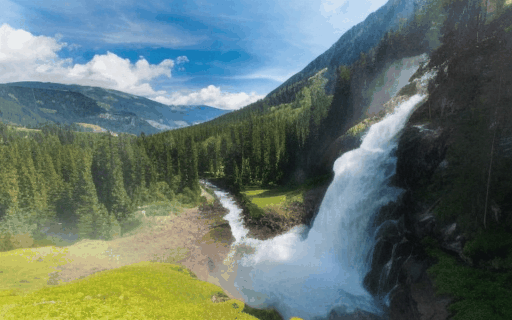} &
        \includegraphics[width=.25\linewidth]{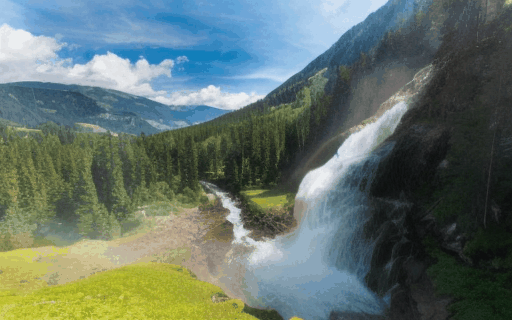} &
        \includegraphics[width=.25\linewidth]{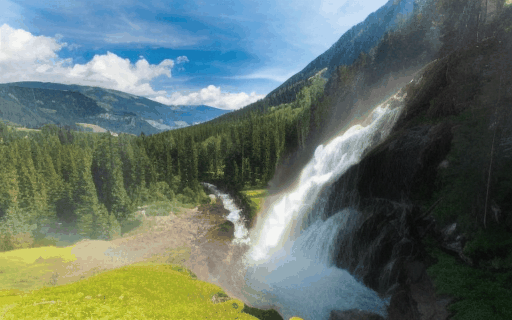}\\
        Input+Drag & Frame 2 & Frame 9 & Frame 16 \\
    \end{tabular}
    \end{center}
    \caption{\textbf{Combination of motion trajectories and motion brush}. Motion-I2V supports the combined usage of motion brush and trajectory guidance.}
    \label{fig:motion-brush_and_drag}
\end{figure}

Our framework also naturally supports region-specific I2V, where only user-specified regions of the reference image are animated, as shown in Fig.~\ref{fig:motion-brush}. It also can be used in combination with the sparse trajectory guidance, as shown in Fig.~\ref{fig:motion-brush_and_drag}, for more controllability.

This is a natural extension of the aforementioned sparse trajectory guided I2V. Specifically, the input $f_{sparse}$ is set to all-zero maps. As for the mask $m$, user-specified regions are set to $0$ and other areas as $1$. Intuitively, this setting requires the ControlNet to keep the non-specified regions static while infer plausible motions for the user-specified regions.


\subsection{Zero-Shot Video-to-Video Translation}

\begin{figure}[h]
    \small
    \setlength{\tabcolsep}{0.75pt}
    \begin{center}
    \begin{tabular}{@{} c c c c @{}}
        \includegraphics[width=.25\linewidth]{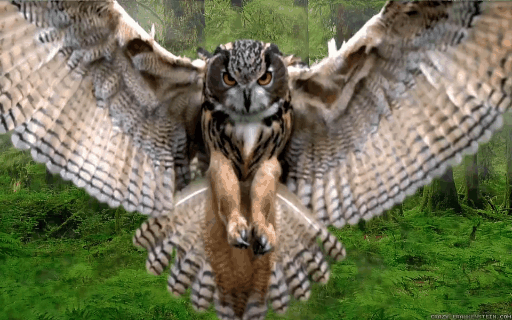} &
        \includegraphics[width=.25\linewidth]{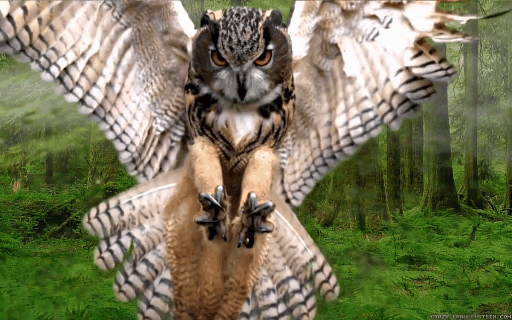} &
        \includegraphics[width=.25\linewidth]{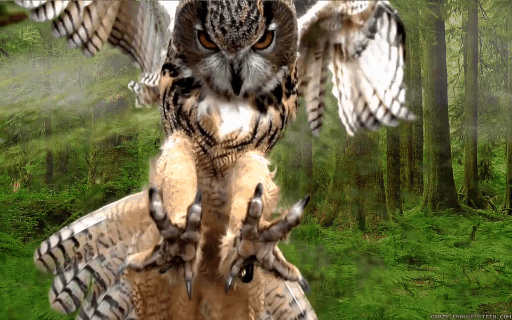} &
        \includegraphics[width=.25\linewidth]{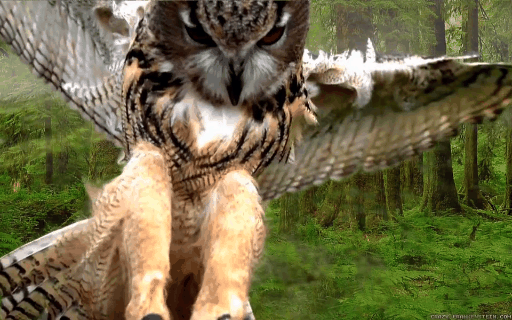}\\
        \includegraphics[width=.25\linewidth]{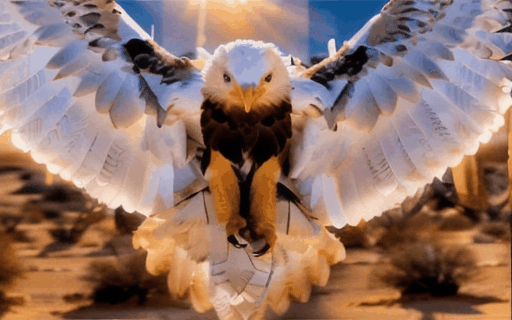} &
        \includegraphics[width=.25\linewidth]{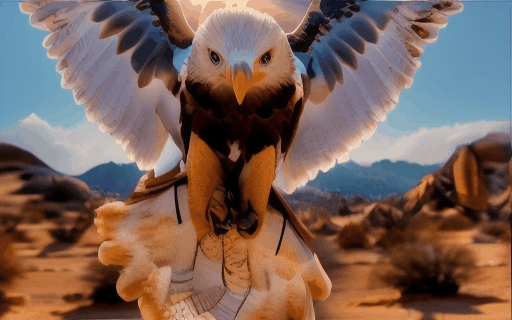} &
        \includegraphics[width=.25\linewidth]{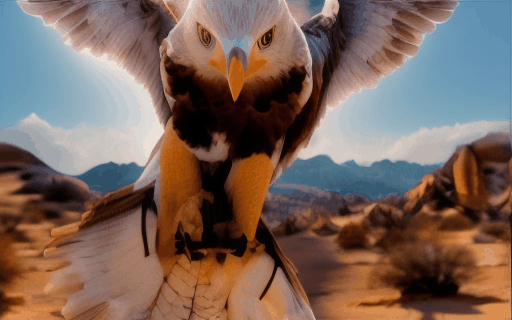} &
        \includegraphics[width=.25\linewidth]{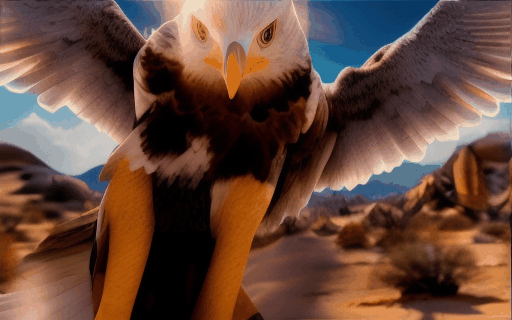}\\
        Frame 1 & Frame 6 & Frame 11 & Frame 16 \\
    \end{tabular}
    \end{center}
    \caption{\textbf{Example of video-to-video translation.} The second stage of Motion-I2V can be used for zero-shot video-to-video translation. The first frame of source video is transformed into the target style using existing image-to-image tools. Then the transformed image can be animated using the second stage of Motion-I2V guided by the motions from the source video.}
    \label{fig:v2v}
\end{figure}

Our framework also naturally supports video-to-video translation (V2V), where a given video is rendered into a new video of another artistic expression specified by the text prompt, as shown in Fig~\ref{fig:v2v}. Specifically, users can utilize existing image-to-image tools to convert the first frame into the target style. Then the displacement fields of the source video can be predicted using off-the-shelf dense point tracker and are used to animate the converted first frame with the second stage of our Motion-I2V.

\begin{figure*}[ht]
    \small
    \setlength{\tabcolsep}{1.25pt}
    \begin{center}
    \begin{tabular}{@{} c c c c c c @{}}
         &
        \includegraphics[width=.195\linewidth]{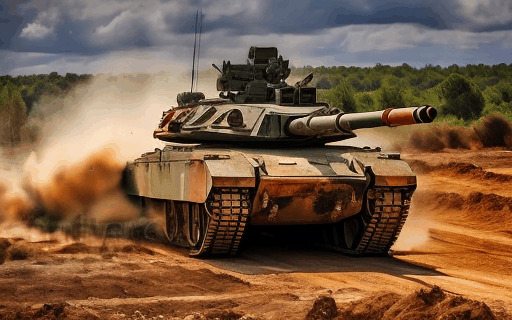} &
        \includegraphics[width=.195\linewidth]{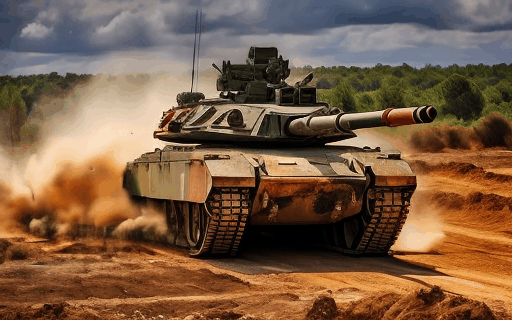} &
        \includegraphics[width=.195\linewidth]{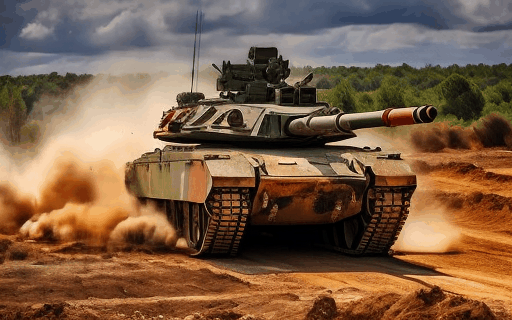} &
        \includegraphics[width=.195\linewidth]{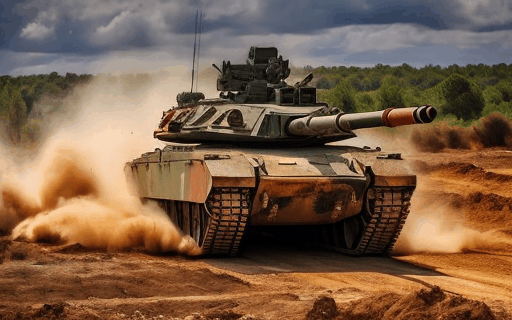} 
        & \rotatebox{270}{\hspace{-55pt}DynamiCrafter} \\
        \includegraphics[width=.195\linewidth]{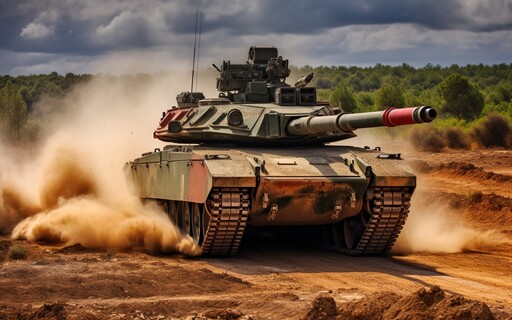} &
        \includegraphics[width=.195\linewidth]{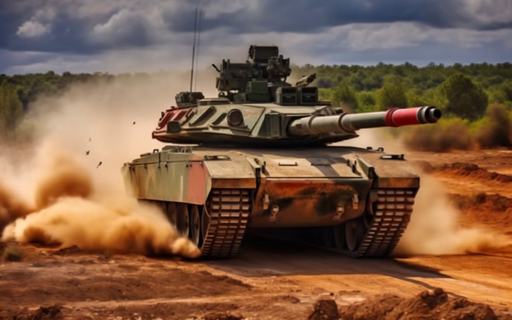} &
        \includegraphics[width=.195\linewidth]{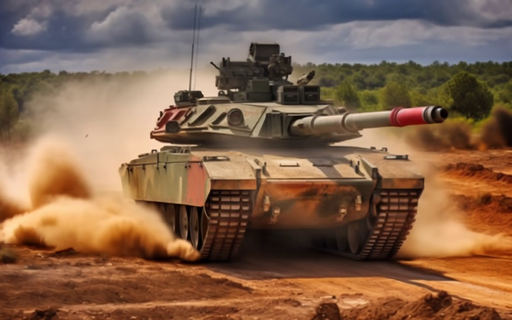} &
        \includegraphics[width=.195\linewidth]{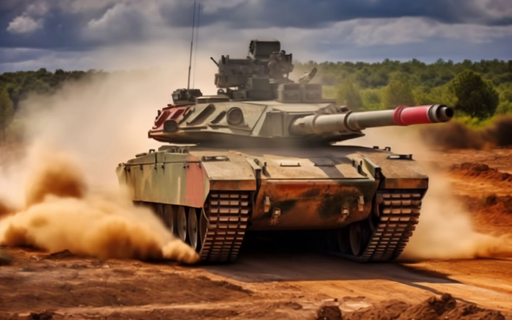} &
        \includegraphics[width=.195\linewidth]{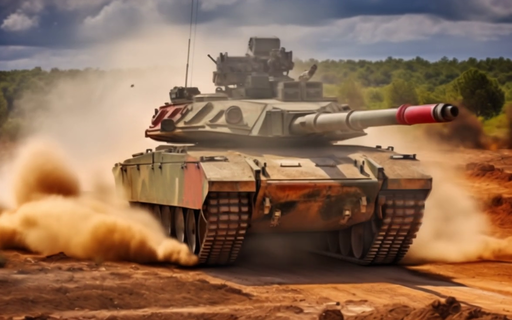}
        & \rotatebox{270}{\hspace{-40pt}Pika 1.0} \\
        A fast driving tank. &
        \includegraphics[width=.195\linewidth]{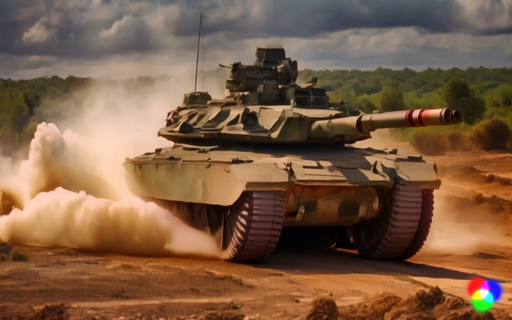} &
        \includegraphics[width=.195\linewidth]{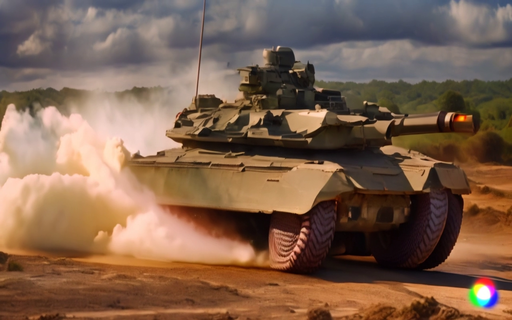} &
        \includegraphics[width=.195\linewidth]{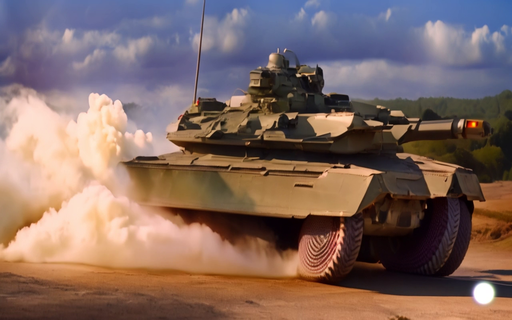} &
        \includegraphics[width=.195\linewidth]{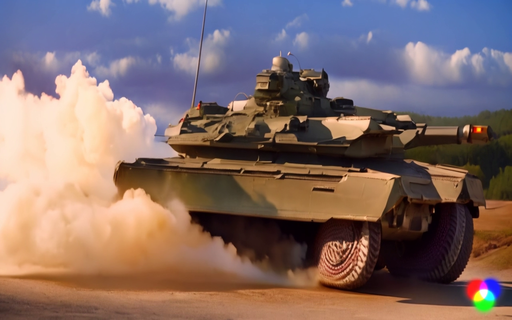}
        & \rotatebox{270}{\hspace{-40pt}Gen2} \\
        &        \includegraphics[width=.195\linewidth]{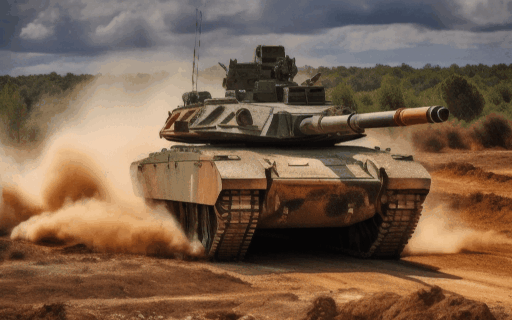} &
        \includegraphics[width=.195\linewidth]{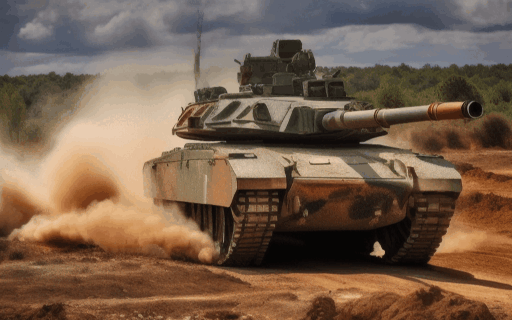} &
        \includegraphics[width=.195\linewidth]{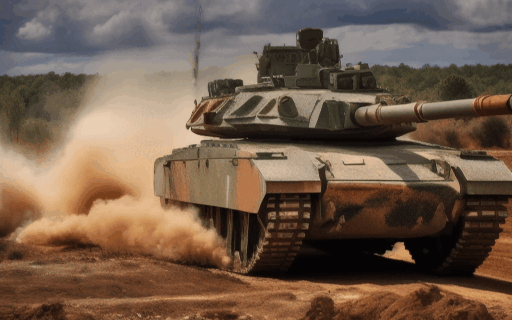} &
        \includegraphics[width=.195\linewidth]{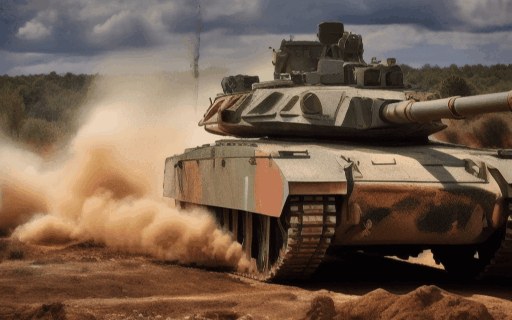} 
        & \rotatebox{270}{\hspace{-40pt}\textbf{Ours}} \\
         &
        \includegraphics[width=.195\linewidth]{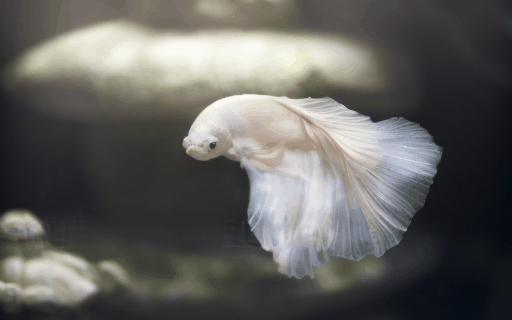} &
        \includegraphics[width=.195\linewidth]{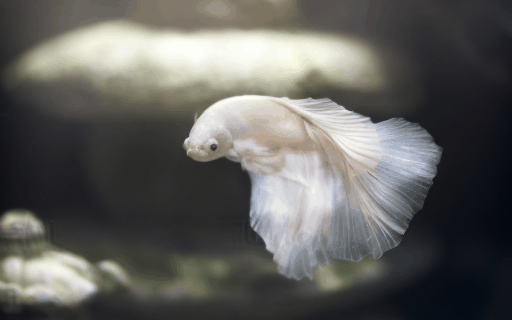} &
        \includegraphics[width=.195\linewidth]{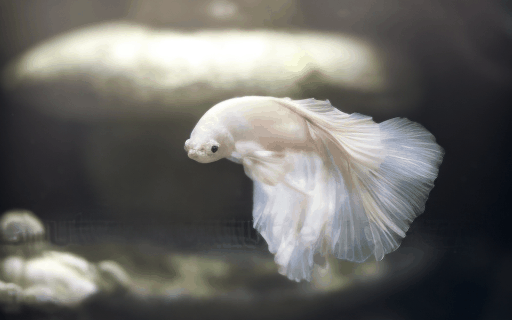} &
        \includegraphics[width=.195\linewidth]{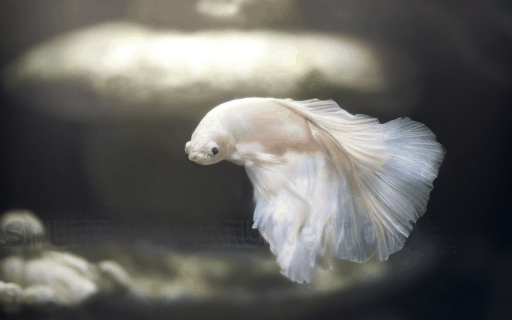}
        & \rotatebox{270}{\hspace{-55pt}DynamiCrafter} \\
        \includegraphics[width=.195\linewidth]{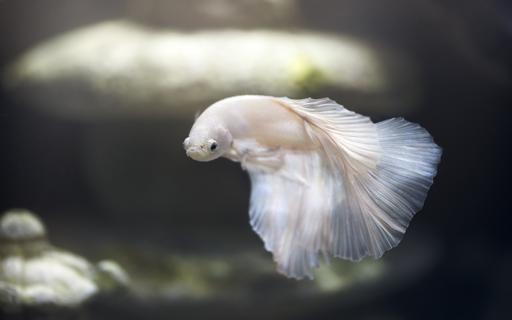} &
        \includegraphics[width=.195\linewidth]{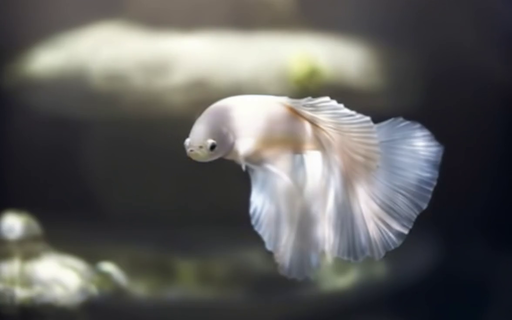} &
        \includegraphics[width=.195\linewidth]{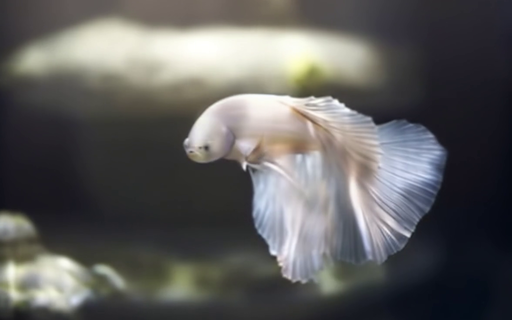} &
        \includegraphics[width=.195\linewidth]{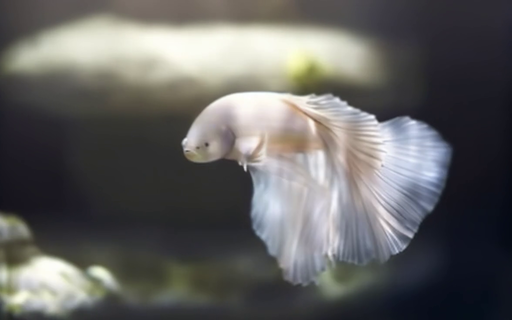} &
        \includegraphics[width=.195\linewidth]{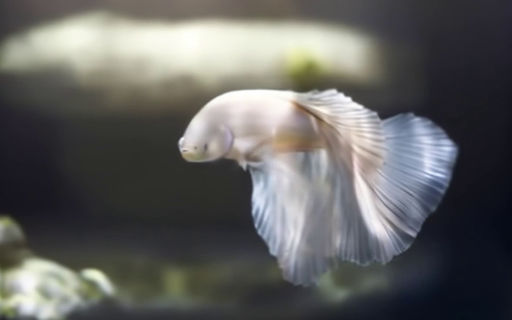}
        & \rotatebox{270}{\hspace{-40pt}Pika 1.0} \\
        A swimming white Betta fish. &
        \includegraphics[width=.195\linewidth]{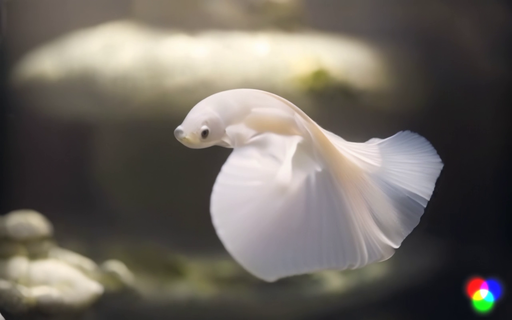} &
        \includegraphics[width=.195\linewidth]{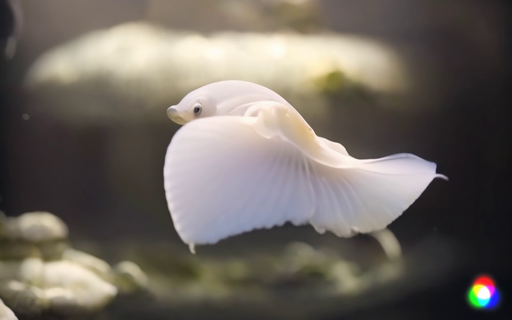} &
        \includegraphics[width=.195\linewidth]{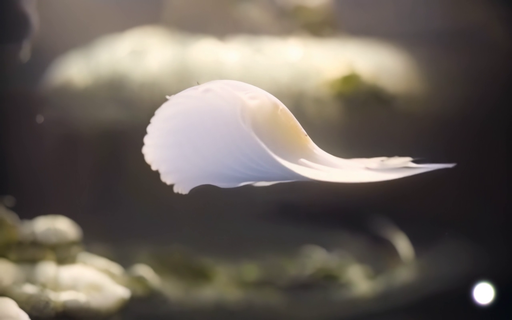} &
        \includegraphics[width=.195\linewidth]{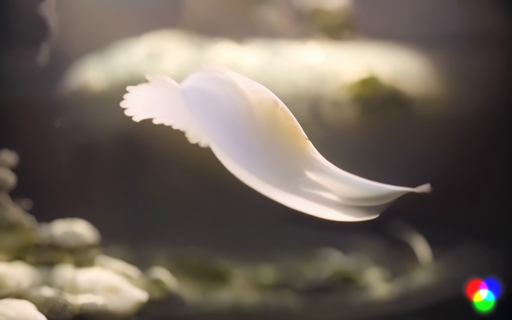}
        & \rotatebox{270}{\hspace{-40pt}Gen2} \\
         &
        \includegraphics[width=.195\linewidth]{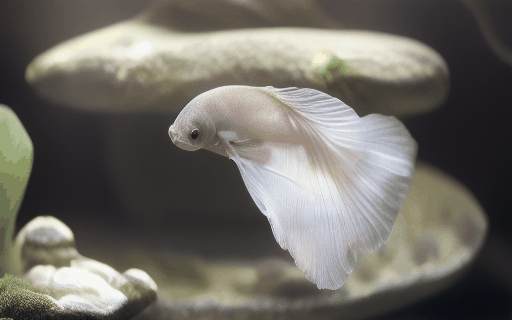} &
        \includegraphics[width=.195\linewidth]{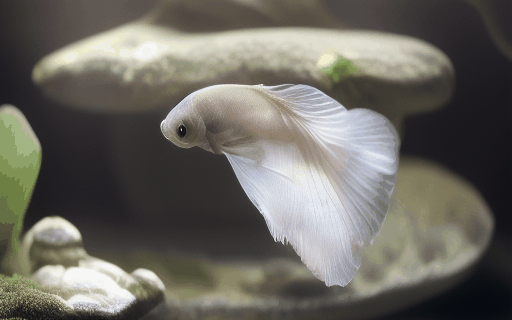} &
        \includegraphics[width=.195\linewidth]{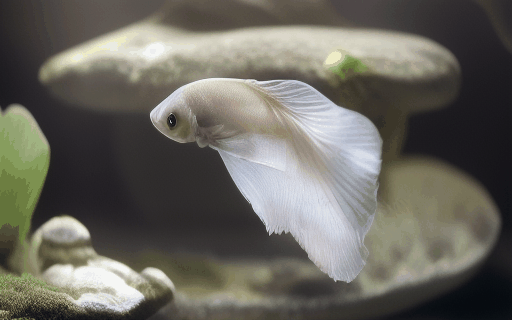} &
        \includegraphics[width=.195\linewidth]{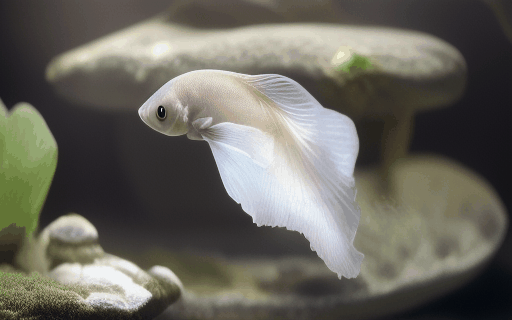}
        & \rotatebox{270}{\hspace{-40pt}\textbf{Ours}} \\
    \end{tabular}
    \end{center}
    \caption{\textbf{Qualitative comparison.} DynamiCrafter and Pika 1.0 tend to generate videos of very small motions. Gen2 can generate as large motion as our method, but fails to preserve the identity of the reference image. Our Motion-I2V can synthesize temporally consistent videos in the presence of large motions.}
    \label{fig:qualitative}
\end{figure*}

\section{Experiments}
\subsection{Experimental Setup}
\noindent \textbf{Training.} We choose Stable Diffusion v1.5 as the base LDM model for the first stage and AnimateDiff v2 as the base VLDM for the second stage. All models are trained on WebVid-10M~\cite{Bain21}, a large scale text-video dataset. During training, we randomly sample 16-frame video clips with a stride of 8. We employ the AdamW~\cite{loshchilov2017decoupled} optimizer with a constant learning rate of $3\times10^{-5}$ for training all models. Stage 1 is trained with videos of resolution of $320 \times 512$ and stage 2 is $320 \times 320$. All experiments are conducted on 32 NVIDIA A100 GPU. Please refer to supplementary for more training details.

\noindent \textbf{Evaluation.} There are a few image-to-video benchmarks, but they are limited to specific domains. For extensive evaluation, we build a test set that covers various categories, such as human activity, animals, vehicle, natural scenes and AI-generated images. It contains $80$ images downloaded from the copyright-free website Pixabay. We use ChatGPT-4V to generate prompts for the image content and possible motion. We use CLIP text-image logits to measure the prompt consistency between prompt and generated frames. We calculate the cosine similarity between consecutive generated frames in the CLIP embedding space to measure the temporal consistency. We further estimate the optical flows between the first and subsequent generated frames to show the motion magnitude.

\subsection{Comparison with Other Methods}

For quantitative evaluation, we compare our method with the open-sourced state-of-the-art methods VideoComposer~\cite{2023videocomposer}, I2VGen-XL~\cite{2023i2vgenxl} and DynamiCrafter~\cite{xing2023dynamicrafter}. Detailed results are shown in Table.~\ref{Tab:quantitative.}. Our Motion-I2V outperforms other methods in the prompt-following metric. Additionally, Motion-I2V can generate more consistent videos even with larger motions.

For qualitative comparison, due to limited space, we compare with the quantitatively second best method DynamiCrafter, together with two commercial products Pika1.0 and Gen2. Results are shown in Fig.~\ref{fig:qualitative}. We observe that DynamiCrafter is not sensitive to motion prompts and tends to generate videos with small motion. This observation is in line with the quantitative results. Pika 1.0 shares similar limited motions but offers better visual quality. Gen2 can generate motions as large as those produced by Motion-I2V, but it suffers from severe distortion. These results verify that Motion-I2V has superior performance in generating consistent results even at the presence of large motions.

\setlength{\tabcolsep}{1pt}
\begin{table}[h]
\centering
\scriptsize
\begin{tabular}{cccc}
\hline
Method & Prompt Consistency $\uparrow$ & Frame Consistency $\uparrow$ & Average Displacement\\

\hline
 VideoComposer & 32.62 & 0.9393 & 67.15 \\
 I2VGen-XL & 33.69 & 0.9650 & 17.70\\ 
 DynamiCrafter & 34.60 & 0.9860 & 3.31\\
 Ours & 34.86 & 0.9871 & 20.06 \\
 
\hline \\
\end{tabular}
\caption{\label{Tab:quantitative.} $\textbf{Quantitative comparison.}$ Motion-I2V shows best instruction-following ability and temporal consistency. Meanwhile, Motion-I2V generates relatively large motions.
}
\end{table}

\subsection{Ablation Study}

We conduct ablation studies to evaluate the effects of critical design choices. We first train a model without stage 1 (first row of Table.~\ref{Tab:ablation.}) where temporal dependencies are solely learned by the 1-D temporal module. We observe that this model is unstable during inference and easy to generate crashed results. This is in line with the low consistency scores and extremely large motions. Then we add stage 1 but utilize the predicted motion fields in a naive way: directly adding the warped feature maps $z[i]'$ to $z[i]$ rather than using attention to adaptively inject the warped feature to subsequent frames. As shown in the second row of Table.~\ref{Tab:ablation.}, this additional motion information stabilizes the prediction, leading to higher consistency score and more vivid motions. By further changing the fusion type to attention as Equation~\ref{soft-inject}, we obtain the final model that achieves the highest consistency score.

\setlength{\tabcolsep}{1pt}
\begin{table}[h]
\centering
\scriptsize
\begin{tabular}{ccccc}
\hline
Stage 1?& Fusion Type & Prompt Consis. $\uparrow$ & Frame Consis. $\uparrow$ & Average Displacement\\

\hline
\xmark & - & 32.95 & 0.9505 & 66.44 \\
\cmark & Addition & 33.99 & 0.9542 & 48.91\\ 
\cmark & Attention & 34.86 & 0.9871 & 20.06\\
 
\hline \\
\end{tabular}
\caption{\label{Tab:ablation.} $\textbf{Ablation study.}$ Utilizing the motion fields from stage 1 can significantly stabilize the prediction. Additionally, using attention to adaptively inject the warped features into synthesized frames can further increase consistency and avoid extreme distortions.
}
\end{table}

\section{Limitations and Conclusions}

We observe that our method tends to generate videos of medium brightness. This is likely because the noise schedule does not enforce the last timestep to have zero signal-to-noise (SNR) ratio as discussed in~\cite{lin2024common}. This flawed schedule leads to training-test discrepancy and limit the model's generalization. We believe using the latest Zero-SNR schedulers can alleviate this issue. To conclude, in this paper we propose a novel I2V framework that factorizes the difficult image to video generation task into two stages. In the first stage, we train a diffusion-based motion fields predictor that focuses on deducing the plausible motions. It shows great motion generative capacity. In the second stage of video rendering, we identify that the naive 1-D temporal attention limits the temporal modeling capacity. To effectively enlarge temporal receptive field, we propose the motion-guided temporal attention. We further explore to provide more controls over the I2V generation process by training a ControlNet for the first stage. We believe controllability of I2V will obtain more attention from the community in the future.

\section{Acknowledgements}
This project is funded in part by National Key R\&D Program of China Project 2022ZD0161100, and in part by General Research Fund of Hong Kong RGC Project 14204021.

{\small
\bibliographystyle{ieee_fullname}
\bibliography{egbib}
}

\end{document}